%% file: main.tex
\newif\ifnips\nipstrue
\newif\iffinalversion\finalversiontrue

\documentclass{article} % For LaTeX2e
\usepackage{iclr2021_conference,times}

\input{math_commands.tex}

\usepackage{hyperref}
\usepackage{url}

\iffinalversion\iclrfinalcopy\fi

\usepackage[utf8]{inputenc} % allow utf-8 input
\usepackage[T1]{fontenc}    % use 8-bit T1 fonts
\usepackage{hyperref}       % hyperlinks
\usepackage{booktabs}       % professional-quality tables
\usepackage{amsfonts}       % blackboard math symbols
\usepackage{nicefrac}       % compact symbols for 1/2, etc.
\usepackage{microtype}      % microtypography

\usepackage{xspace,wrapfig,amstext,marvosym}
\usepackage[hide]{ed}

\usepackage{paralist}

\newcommand{\mmt}{\texorpdfstring{{\normalfont\scshape{Mmt}}\xspace}{MMT\ }}
\newcommand{\omdoc}{{\scshape{OMDoc}}\xspace}
\let\oldlatex\LaTeX
\newcommand{\oldstex}{\texttt{s}\TeX}
\newcommand{\sTeX}{\oldstex\xspace}
\renewcommand{\LaTeX}{\oldlatex\xspace}

\usepackage{tikz}%,stex,amstext,amssymb}
\usepackage{verbatim,listings,amsthm}

\title{Disambiguating Symbolic Expressions\\ in Informal Documents}
\author{%
  Dennis Müller\\% \thanks{Use footnote for providing further information about author (webpage, alternative address)---\emph{not} for acknowledging funding agencies.} \\
  Knowledge Representation and Management\\
  FAU Erlangen-Nürnberg\\[2mm]
  Computational Logic\\
  University of Innsbruck\\
  \texttt{d.mueller@kwarc.info}
  \And
  Cezary Kaliszyk\\
  Computational Logic\\
  University of Innsbruck\\[2mm]
  Institute of Computer science\\
  Warsaw University\\
  \texttt{cezary.kaliszyk@uibk.ac.at}
}

\usetikzlibrary{arrows}
\usetikzlibrary{decorations.pathreplacing}

\newtheorem{definition}{Definition}[section]
\newtheorem{example}{Example}[section]

\begin{document}
\maketitle

\begin{abstract}
	We propose the task of \emph{disambiguating} symbolic expressions in informal STEM documents in the form of \LaTeX files -- that is, determining their precise semantics and abstract syntax tree -- as a neural machine translation task. We discuss the distinct challenges involved and present a dataset with roughly 33,000 entries. We evaluated several baseline models on this dataset, which failed to yield even syntactically valid \LaTeX before overfitting. Consequently, we describe a methodology using a \emph{transformer} language model pre-trained on sources obtained from \url{arxiv.org}, which yields promising results despite the small size of the dataset. We evaluate our model using a plurality of dedicated techniques, taking the syntax and semantics of symbolic expressions into account.
\end{abstract}

\input{intro}

\edef\lquot{``}
\lstMakeShortInline[basicstyle={\small\ttfamily}]`

\input{prelim}

\input{sota}

\input{translation}

\lstDeleteShortInline`

\input{conc}

\iffinalversion
\section*{Acknowledgments}
The first author and this work were supported by a postdoc fellowship of 
the German Academic Exchange Service (DAAD).

The second author is supported by ERC starting grant 
no.\ 714034 \textit{SMART}
\fi

\newpage

\input{main.bblx}
\clearpage
\appendix
\input{appendices}
\end{document}

%% file: math_commands.tex
\usepackage{amsmath,amsfonts,bm}

\def\eqref#1{equation~\ref{#1}}

\def\1{\bm{1}}

\DeclareMathAlphabet{\mathsfit}{\encodingdefault}{\sfdefault}{m}{sl}
\SetMathAlphabet{\mathsfit}{bold}{\encodingdefault}{\sfdefault}{bx}{n}

%% file: intro.tex
\section{Introduction}\label{sec:intro}

Despite huge advancements in machine learning, the task of understanding informal reasoning is still beyond current methods. In fact, it became
commonplace that humans annotate informal documents containing reasoning in many domains, e.g. law~\citep{LibalS20}. Reasoning is most visible in
mathematical documents and software specification and as such in the last decades, the formalization of mathematical knowledge, and the
verification
of formal proofs, has become increasingly popular. By now, dozens of interactive and automated theorem prover systems
are available, each providing libraries with up to hundreds of thousands of formalizations of mathematical definitions, theorems, and their proofs
written by human mathematicians~\citep{HarrisonUW14}.

While formal methods are still primarily used by computer scientists (e.g. to verify software and hardware, as well as in program synthesis), by now they have also drawn the interest of an increasing number of research mathematicians -- primarily thanks to famous problems such as Kepler's conjecture \citep{HalAdaBau:afpkc15} or the classification theorem for finite simple groups \citep{Solomon:fsgc95}, which have successfully been verified using theorem prover systems.

However, while \emph{some} mathematicians have begun actively adapting formal methods for their work, there is a prohibitively large discrepancy between the way new mathematical results are developed, presented, and published in mathematical practice, and the way they are formalized and implemented in formal systems~\citep{kaliszyk2020survey}: Most theorem proving systems implement a fixed \emph{logical foundation} (such as variants of set theory or various kinds of type theories), a surface syntax in which a user declares new definitions and statements in terms of the underlying foundations, and either a tactic language or a language for expressing \emph{proof terms} (usually on basis of the Curry-Howard-correspondence in a typed $\lambda$-calculus) that allow for declaring proofs. Consequently, the process of formalizing new content in a formal system resembles \emph{programming} much more than it does developing informal proofs. % Taking Thomas Hales' proof of Kepler's conjecture as an example, the original informal proof was announced in 1998; its full formalization was completed a full 16 year later in 2014, using a combination of two proof assistants (HOL Light~\citep{hollight} and Isabelle~\citep{isabelle}), requiring additional development on these systems and of new proof tactics and was finally published as the joint work of 22 authors. Overall, it is estimated that this effort of going from a fully developed informal proof -- as accepted by the mathematical community --, to a fully formal proof verifiable by a proof assistant took up to 30 person years.

This discrepancy results in severe challenges for traditional mathematicians:
Formal systems are difficult to learn and use, even if one is well acquainted with the (informal) mathematics involved.
They require learning dedicated formal languages resembling programming languages, declaring content on a level of detail that is prohibitive for beginners even for ``obvious'' conclusions,
	%\item 
	and their libraries are difficult to grasp without already being familiar with the system's language, conventions and functionalities. Due to the required level of detail, knowledge of the existing libraries is crucial when formalizing new content. Furthermore, many ``intuitively valid'' arguments can not be easily expressed in terms of a logical foundation in the first place, and knowing how to deal with those requires familiarity with the logical foundation involved and lots of practice. % (See  for a survey deatilling some of these aspects) % ().

Consequently, the utility of formalizing mathematical results can be too easily (and too often \emph{is}) dismissed in light of the additional time and work required for non-experts. This is despite the fact that many services available for formal mathematics are already enabled by \emph{semi-}formal (or \emph{flexiformal}) representations, such as semantic annotations in natural language texts, or formal representations containing opaque informal expressions (see e.g. \citet{Kohlhase:tffm13,Lange:PhD,Iancu:phd,KohKopMue:mmrdftg17,CS2017,DehKohKon:iop16}). Therefore, we need to invest into methods for bridging the gap between informal mathematical practice and (semi-)formal mathematics. One way to do so is to investigate \emph{autoformalization}, the task of (semi-automatically) converting existing informal mathematical presentations to (increasingly) formal representations.

Notably, these issues extend beyond pure mathematics to other STEM (science, technology, engineering and math) fields, where the formal verification (or lack thereof) of results can have direct real-world implications -- examples include an infamous and costly error in the floating-point unit of Intel processors~\citep{1210044} and several human failures to adequately convert between SI and imperial units, most famously in NASA's Mars orbiter~\citep{marsorbiter:on}. In fact, the former has already established formal verification as a vital tool in hardware design~\citep{1210044}. %, the latter has not yet established the importance of formal methods in other areas.

Two observations motivate the research presented here:
\begin{enumerate}
	\item The vast majority of STEM researchers can be assumed to be comfortable with using \LaTeX; any integration of formal methods in a \LaTeX\ development environment (e.g. via new packages or IDE integration) would consequently lower the entry barrier significantly.
	\item The task of going from purely informal mathematical texts to fully formal representations of the contained knowledge is best done via a separation of concerns, by focussing on individual subtasks (such as disambiguating symbolic expressions, parsing natural language, and translating it to a formal foundation) using dedicated tools for each.
\end{enumerate}

In this paper, we discuss specifically the task of \emph{disambiguating} symbolic expressions -- i.e. associating all symbols in an expression with their precise semantics -- in \LaTeX\ documents as a machine learning task, using \sTeX semantically annotated \LaTeX{} \citep{stex}. The contributions are threefold:

\begin{inparaenum}
	\item We discuss the details of disambiguating symbolic expressions in informal STEM documents as a neural machine translation task,
	\item we present a new dataset specifically for this task, based on the existing SMGLoM library of \sTeX macros (see \autoref{sec:smglom}), and
	\item we present a methodology (using transformer language models) that allows us to achieve positive results on our dataset. We previously evaluated several baseline NMT models (such as \citet{luong17,DBLP:journals/corr/VaswaniSPUJGKP17} and a plain character-based sequence-to-sequence model), which all failed to yield meaningful results due to our dataset being considerably smaller than is required for traditional NMT models.\footnote{All code and data relevant to this paper is available at \url{https://gl.kwarc.info/dmueller/fifom}.}
\end{inparaenum}

%% file: prelim.tex
\section{Preliminaries}

By \emph{disambiguating}, we mean the task of transforming a sequence of symbols
(representing a ma\-the\-ma\-ti\-cal formula) into an \emph{abstract syntax tree} and
associating each leaf in the tree with a unique identifier specifying the precise
semantics of the corresponding symbol.

While this might superficially seem an easy task, closer consideration shows that
even obvious seeming statements such as \lquot$a+b$'' can in fact correspond to a
multitude of possible disambiguations: $a$ and $b$ can be variables or previously
defined constants, whereas $+$ can represent e.g. addition on multiple different
number spaces, generic ring or vector space operations, or string concatenation.
In order to adequately disambiguate expressions generically, it is, therefore, necessary to
take the context in which the expression occurs into account. 

In this paper, we consider informal documents in \LaTeX\ specifically, which we
will disambiguate with the \emph{\sTeX} package, using semantic identifiers provided
by the \emph{SMGloM} library. This eventually enables various formal knowledge management services
(such as type/proof checking) provided by the \mmt system.

\subsection{sTeX}

Kohlhase proposed \sTeX\citep{stex}, a package for annotating \LaTeX documents with structural and formal semantics which is today used by multiple groups formalizing mathematics in various systems.
In particular, \sTeX is based on \omdoc~\citep{omdoc}, an extension of OpenMath~\citep{openmath} which is foundation-agnostic in the sense that it does not favor a specific foundation (such as type or set theories) over any other. This approach is consequently best suited for semantifying informal documents, where foundations are often unspecified, left implicit or switched fluently. For example, category-theoretic and set-theoretic formulations are often used interchangeably in algebraic settings, whereas type theories are generally favored for computational aspects and formal systems. 

\autoref{ex:stex} shows example \sTeX macros and their usage in various stages. Relevant for this paper is primarily the `\symdef` command, which introduces a new mathematical concept (e.g. `\nattimes` in \autoref{ex:stex}). It takes as arguments a macro name (e.g. `nattimes`), a symbolic notation (last argument) and optionally an \omdoc-name (e.g. `multiplication`), arity (e.g. `[1]`, which may be flexary) and notational precedence (e.g. `p=600`, for automatic bracketing). It generates a unique identifier for the concept being declared (based on the provided \omdoc-name), and a new \LaTeX macro (e.g. `\nattimes`) for referring to %(applications of)
the symbol.
Alternative notational variants for symbols can be introduced via `\symvariant`, which are used as options to the macro (e.g. `\nattimes[cdot]`).

In addition to being valid \LaTeX, compilable via \texttt{pdflatex}, \sTeX-documents can be transformed to \omdoc using the LaTeXML-software~\citep{GinevSMK11}, yielding a formally disambiguated representation of the document and in particular the symbolic expressions therein on the basis of the macros provided by `\symdef`s. LaTeXML also heuristically attempts to disambiguate non-\sTeX-symbols, e.g. by considering \lquot='' and \lquot+'' as infix notations for generic equality and addition operators, respectively. 

\begin{figure}[t!]
\begin{center}\begin{tabular}{|l|c|}\hline
 \begin{minipage}[t]{0.23\textwidth}\sTeX declarations\\(signature module) \end{minipage} & \begin{minipage}[t]{0.67\textwidth}\small
  `\begin{modsig}{natarith}`\\[-1mm]
  `  ...`\\[-.5mm]
  `  \symdef[name=multiplication]{nattimesOp}{\*}`\\
  `  \symvariant{nattimesOp}{cdot}{\mathop\cdot}`\\%` `\\[-2mm]
  `  \symdef[assocarg=1,name=multiplication]`
  `    {nattimes}[1]{\assoc[p=600]{\nattimesOp}{#1}}`\\
  `  \symvariant{nattimes}[1]{cdot}`\\
  `    {\assoc[p=600]{\nattimesOp[cdot]}{#1}}`\\
  `...`
  `\end{modsig}`\\[-1mm]` `
  	\end{minipage} \\\hline
  	\begin{minipage}[t]{0.23\textwidth}\sTeX references\\(natural language\\ module) \end{minipage}
   & \begin{minipage}[t]{0.68\textwidth}\small
`\begin{mhmodnl}{natarith}{en}`\\[-1mm]
`  ...`\\[-.5mm]
`  \begin{definition}`\\
`    \Defi{multiplication} $\nattimesOp[cdot]$`
`    computes the \defi{product} $\nattimes[cdot]`
`      {a,b}$ (also written as $\nattimes{a,b}$ or`
`    $\nattimes[x]{a,b}$) of \trefiis[naturalnumbers]`
`      {natural}{number} $a$ and $b$. It is defined`
`    by the equations $\eq{\nattimes[cdot]{x,0},0}$`
`    and $\eq{\nattimes[cdot]{x,\natsucc{y}},`\\
`     \natplus{x,\nattimes[cdot]{x,y}}}$.`\\
`  \end{definition}`\\
`...`
`\end{mhmodnl}`\\[-1mm]` `
  	\end{minipage}\\\hline
  	\begin{minipage}[t]{0.23\textwidth}PDF output\\(for the natural\\ language module) \end{minipage}
  & \begin{minipage}[t]{0.67\textwidth}\small
  	\textbf{Definition. }\textcolor{red}{Multiplication} $\cdot$ computes the \textcolor{red}{product}
$a\cdot b$ (also written as $a\*b$ or $a\times b$) of \textcolor{blue}{natural numbers}
$a$ and $b$. It is defined by the equations $x\cdot0=0$ and $x\cdot S(y)=x+x\cdot y$.
  \end{minipage}\\\hline
  \omdoc & \begin{minipage}[t]{0.67\textwidth}\small
  	`<OMA>`\\
  	`  <OMS cd="smglom:mv?equal" name="equal"/>`\\
  	`  <OMA>`\\
  	`    <OMS cd="smglom:arithmetics?natarith"`\\
  	`        name="multiplication"/>`\\
  	`    <OMV name="x"/>`\\
  	`    <OMI>0</OMI>`\\
  	`  </OMA>`\\
  	`  <OMI>0</OMI>`\\
  	`</OMA>`\\[-2mm]
  \end{minipage}\\\hline
\end{tabular}\vspace{-3mm}
\end{center}
\caption{\label{ex:stex} An \sTeX Example: The \omdoc corresponds to the symbolic expression $x\cdot 0=0$}
\end{figure}

\subsection{SMGLoM}\label{sec:smglom}

\paragraph{} % As a first approach, we will use 
The \emph{SMGloM}~\citep{Kohlhase:dmesmgm14}, \emph{semantic multilingual glossary of mathematics}) is a library of hundreds of \sTeX-modules containing mathematical concepts and definitions. It is separated into \emph{signature modules} (using the `modsig`-environment, see \autoref{ex:stex}) containing only symbol declarations, and \emph{natural language modules} (using the `mhmodnl`-environment, here exemplary for English) that serve as dictionary entries for these, in which the semantics of the symbols are described in a semi-formal manner. %\autoref{ex:smglom} shows an exemplary SMGLoM-entry.
The second row of \autoref{ex:stex} shows an SMGLoM entry.

\subsection{\mmt}\label{sec:MMT}

\sTeX itself is integrated, and shares an underlying \omdoc ontology, with the \mmt system \citep{RK:mmt:10,HKR:extending:12,rabe:howto:14} -- a foundation-independent meta-framework and API for knowledge management services. This integration makes the generic services provided by \mmt -- e.g. type checking, library management/browsing, translation -- available to informal mathematical texts. Using \emph{alignments} \citep{DMueller:phd:19,MueGauKal:cacfms17}, \omdoc-expressions can be translated between different libraries, languages and foundations. This allows for e.g. translating (originally) \sTeX-content to a typed setting in order to e.g. check expressions and run type inference.

Additionally, several theorem prover libraries have been translated to \omdoc and integrated in the \mmt system, e.g. ~\citet{KohMueOwr:mpagsiuf17,MueRabSac:cltg19} (for a detailed overview, see ~\citet{DMueller:phd:19} and ~\citet{KohRab:eempal20}). Extending these integrations to enable exporting from \mmt as well (and in conjunction with natural language processing), this could enable verifying informal mathematics imported via \sTeX using external state-of-the-art theorem prover systems.

%% file: sota.tex
\section{State of the Art}\label{sec:sota}

Various papers over the last years have -- explicitly or implicitly -- 
attempted to extract formal information from informal documents using 
machine learning. These fall into two categories:

Firstly, there are projects that attempt to fully formalize informal 
mathematical documents using machine learning techniques, using the 
surface language of some theorem prover system directly as a target. 
In \citet{cezary1,cezary2,cezary3}, the Flyspeck 
project~\citep{HalAdaBau:afpkc15} -- the formalization of Kepler's 
theorem -- was used as a basis for a parallel dataset in order to 
translate from informal mathematics to HOL Light~\citep{hollight} syntax.
\citet{Kaliszyk2017SystemDS,DBLP:itof,DBLP:conf/cpp/WangBKU20} target the 
Mizar language~\citep{mizar} instead, using the \emph{Journal of 
Formalized Mathematics}~\citep{JFM:base} as data -- an informal 
representation of the formal \emph{Mizar Mathematical 
Library}~\citep{RoleOfMML}.

While these projects achieved impressive results given the ambitious 
nature of the task, their success rate is naturally limited by the 
involved models having to solve several tasks at once 
(see second observation in \autoref{sec:intro}), including ours. 
Additionally, by going to a fully formal language (and logical foundation) 
immediately, the result 
does not preserve the narrative presentation of the input document, 
effectively losing (for us) valuable information in the process.
Consequently, our task and results obtained on it are not directly 
comparable to these projects.

Secondly, various projects have aimed to \emph{solve} informally presented 
mathematical problems of various kinds. These include 
\citet{Arai:2014:MM:2608628.2627488,Matsuzaki:2014:MUE:2893873.2894044,%
matsuzaki-etal-2017-semantic,Matsuzaki2018CanAA} on pre-university math 
problems, \citet{google:neuralmath} and \citet{lample2019deep} on 
high-school level equations, \citet{Gan2017AUTOMATICUA} and 
\citet{seo-etal-2015-solving} on geometric problems, and 
\citet{huang-etal-2018-neural} and \citet{wang-etal-2017-deep} on solving 
typical high-school word problems.

While this naturally entails disambiguating symbolic expressions, all
these projects reduce their domain of applicability to specific
areas where all occurring formal symbols are syntactically unambiguous -- 
primarily common arithmetic operations, functions, and relations on real 
numbers -- such that disambiguation reduces to simple parsing of a fixed,
small set of a priori known symbols.

%% file: translation.tex
\section{Task Definition}

\newcommand{\mathlatex}{\text{\oldlatex}}
\newcommand{\mathstex}{\text{\oldstex}}

\begin{definition} \textbf{(Disamiguation Task)}
	Let $\mathcal L$ be a set of \LaTeX fragments (i.e. strings), which we 
	assume are syntactically valid \LaTeX in \emph{some} suitable document 
	context. 
	
	A \emph{symbolic expression} is (for our purposes, simplified) any 
	substring $s$ of some $S\in L$ such that $s$ is interpreted by the 
	\TeX-engine in \emph{math mode} -- e.g., if it is delimited by 
	\$, \$\$ or `\[` and `\]` respectively.
	
	For the purposes of our task, we call $S\in\mathcal L$ 
	\emph{fully disambiguated}, if every symbolic 
	expression occurring in $S$ only consists of:
	\begin{enumerate}
		\item \emph{variable names} (e.g. `n` or `\mathcal{G}`, provided 
			they do not represent specific, definite mathematical objects),
		\item \sTeX macros introduced via a `\symdef` declaration in the 
			SMGLoM, or
		\item \emph{non-semantic} commands or characters, such as 
			additional spaces/tabs/linebreaks, purely aesthetic spacing or 
			kerning commands, unnecessary parentheses or clarifying 
			comments (e.g. in under- or overbraces).
	\end{enumerate}
	
	Let $\mathcal L_{\mathstex}\subset\mathcal L$ the subset of fully 
	disambiguated \LaTeX fragments. Conversely, let 
	$\mathcal L_\mathlatex\subset\mathcal L$ be the set of \LaTeX 
	fragments that do not contain any \sTeX macros\footnote{Note that 
	$\mathcal L_\mathlatex$ and $\mathcal L_\mathstex$ are not disjoint}.
	
	Clearly, for any $S\in\mathcal L$, there is some 
	$\mathlatex(S)\subset\mathcal L_\mathlatex$ such that $S$ and any 
	$S'\in\mathlatex(S)$ represent the same \emph{symbolic presentation} 
	-- i.e. they generate the same output on `pdflatex`. 
	
	Conversely, we assume that for any $S\in\mathcal L$ there is a set 
	$\mathstex(S)\subset\mathcal L_\mathstex$ such that 
	\begin{inparaenum} \item $\mathlatex(S) = \mathlatex(S')$ for all 
		$S'\in\mathstex(S)$ (i.e. they have the same symbolic presentation) 
		and \item all $S'\in\mathstex(S)$ capture the 
		\emph{intended semantics} of $S$ - i.e. the author of $S$, were 
		they to know the SMGLoM library sufficiently well, would agree 
		that $S'$ is a correctly fully disambiguated variant of 
		$S$.\end{inparaenum}
	
	Our goal is to learn a function $f:\mathcal L\to\mathcal L$ such that 
	for any $S\in\mathcal L$ we have $f(S)\in\mathstex(S)$.
	
\end{definition}

\begin{example}
	Consider the sentence from the SMGloM
	\begin{center}
	`Multiplication $\cdot$ computes the `\ifnips\else\\\fi `product $a\cdot b$ (also written as`
	`$ab$ or $a\times b$) of natural numbers`\ifnips\else\\\fi ` $a$ and $b$.`
	\end{center}

	The last two symbolic expressions (`$a$` and `$b$`) only consist of 
	variable names, and are thus considered \emph{fully disambiguated}
	already.

	The first one (`$\cdot$`) refers to the multiplication operator on
	natural numbers, which in \sTeX is represented as `\nattimesOp`, the
	remaining symbolic expressions are all multiplications
	on natural numbers applied to the variables $a$ and $b$ with different
	notations, represented in \sTeX via `\nattimes` with various options.

	We expect the target function $f$ on this input sentence to output

	\begin{center}
		`Multiplication $\nattimesOp$ computes the product $\nattimes[cdot]{a,b}$` 
		`(also written as $\nattimes{a,b}$ or $\nattimes[x]{a,b}$) of natural` 
		`numbers $a$ and $b$.`
	\end{center}
\end{example}

\section{Datasets}\label{sec:datasets}

We have two datasets of \sTeX-content:
\begin{enumerate}
	\item The SMGLoM\footnote{\url{https://gl.mathhub.info/smglom}}, 
		which introduces precisely those macros that we want to be 
		learned by a model. Unfortunately, it provides relatively few 
		symbols and hence can only cover a small part of informal 
		documents even in theory. Additionally, apart from some 
		rudimentary concepts such as logical 
		connectives or basic arithmetic functions, the SMGLoM library 
		\emph{references} the majority of symbols only once 
		(in the corresponding dictionary entry). This is unlike most other 
		formal systems, where all symbols need to be typed or defined 
		formally when being declared, which naturally leads to a 
		significant number of references to previously 
		declared symbols.
	\item The MiKoMH\footnote{\url{https://gl.mathhub.info/MiKoMH}}-repository 
		of lecture notes by Michael Kohlhase (the author of \sTeX) 
		is heavily biased towards subjects in computer science, covering 
		only a small part of SMGLoM-entries, and often introducing local 
		`\symdef`s.
\end{enumerate}
Notably, while the translation from source to target language is difficult,
the \emph{reverse} translation (from \sTeX to plain \LaTeX) is easy:
Since \sTeX\ macros internally expand (ultimately) to the plain
notational representation as basic \LaTeX, translating from the \emph{target}
to the \emph{source} language amounts to merely expanding \sTeX macros.
This allows for easily generating a parallel dataset from a set of
documents in the target language.

To obtain such a parallel corpus for supervised learning, we take the 
individual \LaTeX-files in those repositories and do the following:
\begin{enumerate}
	\item We separate the documents into small fragments of (on average) 
		500 character lengths, which we consider to be the \emph{sentences} 
		in $\mathcal L_\mathstex$. Symbolic expressions occur preferably at 
		the end of a sentence, based on the assumption that preceding text 
		provides a more meaningful context for disambiguation. Sentences 
		that do not contain symbolic expressions are ignored.
	\item In each sentence $S=S_\mathstex\in\mathcal L_\mathstex$, we 
		perform some standardization function which e.g. removes 
		non-semantic macros and ensures that macro arguments are always 
		braced, in order to minimize author bias,
	\item We extract all symbolic expressions 
		$(m_{\mathstex,i})_{i\leq n_S}$ in $S$ and expand all \sTeX macros 
		in them, resulting in $(m_{\mathlatex,i})_{i\leq n_S}$ 
		(where $n_S$ is the number of symbolic expressions in $S$). 
		Analogously, we expand all \sTeX macros in $S$ itself, yielding 
		$S_\mathlatex\in\mathcal L_\mathlatex$.
\end{enumerate}
Each entry in our dataset then consists of a 4-tuple 
$(S_\mathlatex,S_\mathstex,(m_{\mathlatex,i})_{i\leq n_S},(m_{\mathstex,i})_{i\leq n_S})$. 
In total, we obtain 911 entries from SMGLoM and 9200 entries from MiKoMH.

\paragraph{Synthesizing Training Data}
 In order to augment our datasets for supervised learning, we opted to 
 exploit the \mmt integration to synthesize additional training data.

For that, we aligned SMGLoM symbols with declarations in a strongly typed 
\mmt archive; namely the \emph{Math-in-the-Middle (MitM)} 
library~\citep{DMueller:phd:19}. This allows us to randomly generate 
well-typed (and hence syntactically well-formed) terms in a typed setting, 
translate these along alignments to \sTeX expressions and subsequently 
generate surrounding verbalizations.

The generating algorithm takes as input a set of symbols $\mathtt{Sym}$ 
(e.g. all MitM-symbols for which an alignment to SMGLoM exists) and a 
starting symbol $s\in\mathtt{Sym}$ (e.g. `nattimes`; binary multiplication 
on natural numbers). It returns a random well-typed formal expression 
$t$ which is guaranteed to contain $s$. Afterwards, it is 
\emph{verbalized} as an \sTeX sentence using natural language fragments 
(a detailed description of the algorithm is given in 
\hyperref[apx:synthesizing]{Appendix A}).

The synthesized \sTeX sentences are then treated as above to augment our 
parallel training corpus.

As an \textbf{evaluation dataset}, we developed \sTeX documents based on 
selected fragments of introductory sections from mathematics lecture notes; 
primarily containing basics such as set operations, number spaces, 
examples for proofs by induction, basic combinatorics, and definitions of 
common algebraic structures, containing $161$ symbolic expressions in 
total.  Importantly, these documents were written by hand, with a focus
on featuring multiple symbols with the same symbolic representation;
primarily the usual arithmetic operations on different number spaces.

Of the $\approx100$ SMGLoM symbols used therein, 92 were aligned 
with corresponding symbols in the MitM library and used as input symbols 
for synthesizing sentences; with 250 sentences per starting symbol 
(as to not drown out the non-synthesized sentences), yielding 23,000 
additional sentences.

Unlike the training datasets, the evaluation document was translated to 
plain \LaTeX\ manually using the PDF as a reference, in order to avoid 
possible spurious patterns in automatically expanded \sTeX.

\section{\sTeX-Annotating with Machine Learning as an NMT Task}

In the course of our experiments, we considered our disambiguation task
as a machine translation (NMT) problem, the models for which have been proven 
to be quite effective even beyond natural language 
translations~\citep{clark2020electra}. In fact, the autoformalization 
projects mentiond in \autoref{sec:sota}, which are spiritually closest
to our task, all used NMT models with positive results.
There are however several aspects that distinguish a \LaTeX-to-\sTeX 
translation from similar translation tasks which significantly affect 
the applicability of existing tools and hence our methodology. 

First, 
 Unlike the most popular formal systems, there is no large library of 
 formalizations for the translation target.
This leaves us with only a small dataset that (for the reasons outlined
in \autoref{sec:datasets}) does not represent well the 
general distribution we would like to learn.

Second, translation is only relevant for specific fragments of an input 
text, namely the symbolic expressions; for the surrounding natural 
language texts, translation should be the identity.
	%\item 
Nevertheless, surrounding text usually contains critical information for 
disambiguation; e.g. without the surrounding context, it is impossible to 
disambiguate an expression $a+b$, since the symbol \lquot$+$'' could refer 
to any of dozens of addition operations.

Finally,
 depending on perspective, the domain language is a proper subset of the 
 target language; or rather (since we want to avoid ambiguous expressions 
 in \sTeX) domain and target language share both a basic grammar as well 
 as a large amount of vocabulary 
 (namely $\mathcal L_\mathlatex\cap\mathcal L_\mathstex$) which e.g. 
 subsumes natural English. For the domain language, large datasets are 
 easily obtainable.

Our task could also be considered
as a \emph{text style transfer} task -- e.g. \citet{yang2019unsupervised}
uses pre-trained language models for text style transfer, roughly
similar to (but more sophisticated than) our approach. While the datasets
used therein are still considerably larger than ours, this might be a 
promising avenue for future improvements over our model.

\section{Methodology}

Notably, \sTeX macros reflect the \emph{syntax tree} of an expression,
so that on symbolic expressions \emph{alone}, the representation of the 
target sequences is naturally analogous to those chosen in 
\emph{string-to-tree} translations~\citep{aharoni2017stringtotree}.
Plain \LaTeX however is not naturally amenable to a tree-structured
representation, making \emph{tree-to-tree} 
approaches~\citep{chen2018treetotree} not easily applicable to our
dataset.

Initial experiments using standard, dedicated NMT models with full 
sentences as input/output %using different hyperparameters und tokenizations 
quickly proved to be ineffective due to the size of the training corpus, 
which was too small to cause these models to even generate syntactically 
correct \LaTeX\ (e.g. knowing to balance pairs of brackets) before 
overfitting on the training data. This makes it difficult to compare
our approach to an informative baseline model.

\emph{Transformer language models} (e.g. 
\citet{DBLP:journals/corr/abs-1810-04805,DBLP:journals/corr/abs-1907-11692,Radford2018ImprovingLU,Radford2019LanguageMA,clark2020electra}) 
allow us to leverage huge available corpora of plain \LaTeX\ documents to 
train a model to \lquot understand'' both basic \LaTeX\ syntax and 
mathematical terminology. Using those, we consequently do not need to rely 
on our small dataset for this base-level understanding. We can then 
approach learning \sTeX annotations as a downstream task on a pre-trained 
transformer model.
Consequently, we pre-trained a  GPT2~\citep{Radford2019LanguageMA} model 
on a large portion of available \LaTeX\ sources of scientific papers from 
the preprint repository \url{arxiv.org} 
(6,673,950 entries of length 1,024 tokens). %The documents where normalized as described in \autoref{sec:datasets}. 
The model was trained \emph{from scratch} in order to use a dedicated 
tokenizer trained on \LaTeX\ directly (byte-level tokenizer; vocabulary 
size 32,000) rather than natural language alone.

In order to leverage the pretrained model for both source and target 
language\footnote{Initial experiment with the pretrained model as encoder 
component only showed improvements over randomly initialized 
encoder-decoder-models, but ultimately proved unsuitable still due to the 
small dataset size.}, we subsequently opted to fine-tune the GPT2-model on 
inputs of the form 
\vspace{-2mm}
\begin{center}$S_\mathlatex$` <s> `$m_\mathlatex$` <s> `$m_\mathstex$` <s>`,\end{center}
\vspace{-2mm}
 where `<s>` a single-token separator.\footnote{inspired by \url{http://jalammar.github.io/illustrated-gpt2/\#part-3-beyond-language-modeling}} 
 For example, for \autoref{ex:stex} the training data contains fragments 
 (normalized) such as:
\vspace{-2mm}
\begin{center}
`Multiplication $\cdot$ computes the `\ifnips\else\\\fi `product $a\cdot b$ (also written as`
`$ab$ or $a\times b$) of natural numbers`\ifnips\else\\\fi ` $a$ and $b$.`
`<s> $a\cdot b$ <s> `\ifnips\else\\\fi `$\nattimes[cdot]{a,b}$ <s>`
\end{center}
\vspace{-2mm}
\vspace{-2mm}

We then use text generation on inputs of the form 
$S_\mathlatex$` <s> `$m_\mathlatex$` <s>` for translating and stop 
generating after encountering `<s>`.

By using one entry per symbolic expression, we obtain a dataset of 121,368 
examples. The GPT2-model was finetuned on these for five epochs, resulting 
in an average training loss of 0.04 and yielding promising results on the 
evaluation set (see below). This approach has the following advantages:
\begin{enumerate}
	\item It allows for using large datasets of generic \LaTeX\ documents 
		to learn basic syntactic rules and semantics of mathematical 
		expressions beyond our small \sTeX datasets.
	\item We conjecture that this approach makes the model less sensitive 
		to spurious patterns in the synthesized part of our dataset.
	\item Adding new symbols to the SMGLoM and aligning them to (new or 
		existent) symbols in the MitM library allows for immediately 
		synthesizing training data, obviating the need to first obtain 
		large amounts of data \emph{using} the new symbol before the model 
		can learn to use it.
	\item The mere pretrained GPT2 model can be trained on 
		\emph{additional} downstream tasks, e.g. introducing macros for 
		referencing mathematical concepts in natural language fragments.
\end{enumerate}

\section{Evaluation and Results}

The traditional evaluation metrics (loss during evaluation, perplexity, BLEU) are somewhat difficult and/or meaningless to apply in our situation, since 
\begin{inparaenum}
	\item the returned tokens and provided label tokens might differ in semantically irrelevant ways (e.g. `$a+b$` vs. `$a + b$`), and
	\item loss/perplexity would be evaluated during a forward pass in a next token prediction task on a token-by-token basis, which would retroactively \lquot correct'' errors in prediction that would otherwise yield completely wrong result.
\end{inparaenum}

Consequently, we opted for a plurality of evaluation strategies. Let $S_F$ the returned sentence of our model on an input $S_\mathlatex$ with the correct label $S_\mathstex$. Then on our
 evaluation set 
we get 
\vspace{-2mm}
\begin{enumerate} 
	\item $S_F\in\mathcal L$ for 96.9\% of inputs 
	\item $S_\mathlatex\in\mathlatex(S_F)$ for 64.0\% of inputs, 
	\item $S_F\in\mathcal L_\mathstex$ for 60.2\% of inputs, and 
	\item $S_F = S_\mathstex$ for 47.2\% of inputs.
\end{enumerate}
\vspace{-2mm}
In comparison, using traditional NMT models auch as \citet{luong17,DBLP:journals/corr/VaswaniSPUJGKP17} we effectively obtained 0\% success rates for all of the above.
Additional evaluation techniques exploiting the \mmt integration are described in \hyperref[apx:evaluation]{Appendix B}. 

\begin{figure}[ht!]
\begin{center}\begin{tabular}{rl}
$S_\mathlatex$: & `\mathbb{N}=\{0,1,2,3,\ldots\}` \\
$S_\mathstex$: & `\eq{\NaturalNumbers,\setdots{0,1,2,3}}` \\
$S_F$: & `\NaturalNumbers=\set{0,1,2,3}` \\\hline

$S_\mathlatex$: & `(A \subseteq B)\Leftrightarrow(\forall x\in A. x\in B)` \\
$S_\mathstex$: & `\biimpl{\sseteq{A}{B}}{\foral{\inset{x}{A}}{\inset{x}{B}}}` \\
$S_F$: & `\biimpl{\sseteq{A}{B}}{\foral{x}{A}\inset{x}{B}}}` \\\hline

$S_\mathlatex$: & `\mathcal{P}(A):=\{x|x\subseteq A\}` \\
$S_\mathstex$: & `\defeq{\powerset{A}}{\setst{x}{\sseteq{x}{A}}}` \\
$S_F$: & `\defeq{\powerset{A}}{\bsetst{x}{x}{\sset{x}{x} A}}` \\\hline

$S_\mathlatex$: & `1+2+3+4+5=(5\cdot6)/2=15` \\
$S_\mathstex$: & `\eq{\natplus{1,2,3,4,5},\natdiv[slash]{\nattimes[cdot]`\\
 & `  {5,6}}{2},15}` \\
$S_F$: & `\natplus{1,2,3,4,5}=\natdiv[slash]{\natplus{\nattimes[cdot]`\\
 & `  {5,6},4,5}}{2}=15` \\\hline
\end{tabular}\end{center}
\vspace{-3mm}
\caption{Example Inputs and Outputs from our Evaluation Set}\label{fig:evalex}
\end{figure}

\autoref{fig:evalex} shows a few examples where our model \lquot failed'' in interesting ways. 
As the first and fourth examples show, the model seems to consistently fail to replace \lquot='' by the intended macro `\eq` -- a failure that LaTeXML can recover when converting to \omdoc, but also regularly occurs in the training data. Similarly, `\ldots` often leads to wrong translations: The first example shows that the model simply dropped `\ldots`, using a generic set constructor macro `\set` rather than `\setdots`, the one specifically intended for sets ending in ellipses.

In the second example, the model seems to introduce a nonsensical additional argument for the `\foral` macro. Notably, the expression $\forall x\in A. P$  can also be achieved using the dedicated macro `\foralS{x}{A}{P}`. Seemingly, the model chose the macro `\foral`, and the arguments for the `\foralS` macro, yielding a wrong translation that generates a wrong pdf output, while being \lquot semantically almost correct''.

In the third example, the model confuses the macro `\setst` (for set comprehension) with a more complex macro `\bsetst` (for set comprehension with a complex pattern on the left side). Additionally, it confuses `\sseteq` (for inclusive subsets $x \subseteq A$) with `\sset` (for generic subsets $x\subset A$), duplicating the first argument and moving the \emph{intended} argument `A` outside the scope of the macro.

Example four is interesting in that the model correctly identifies the arithmetic operations as those on the natural numbers, but spuriously inserts an additive term `\natplus{...,4,5}`; this is likely an artifact from the left-hand side of the equation. Interestingly, these kinds of artifacts occur more than once in our evaluation set.

%% file: conc.tex
\section{Conclusion}

We have proposed the task of disambiguating symbolic expressions in informal STEM documents and defined this task formally.
This allows % recognizing if the example expression $a\,b$ in an informal document is supposed to mean the multiplication of the numbers $a$ by $b$ or the concatenation (of two strings in a program), or function application (in functional programming languages), 
for annotating informal documents semantically, and further processing them using tools that support such annotated documents (e.g. \mmt{}).
We discussed the specificity of this task and what separates this task from other NMT problems. We developed a
dataset for this task and presented an approach that yields promising results, especially in light of the size of the dataset.
In particular, the presented approach points to the efficacy of using transformer models pretrained on generic \LaTeX documents.

In the future, we plan to combine the proposed symbolic disambiguation approach with an auto-formalization framework. This way we aim to achieve better results for end-to-end formalization of informal mathematical documents. Furthermore, more promising results for the currently proposed task could be obtained by reintegrating the proposed models into an encoder-decoder NMT model.

%% file: appendices.tex
\section{Synthesizing Training Data}\label{apx:synthesizing}
\lstMakeShortInline[basicstyle={\small\ttfamily}]`

The generating algorithm takes as input a set of symbols $\mathtt{Sym}$ (e.g. all MitM-symbols for which an alignment to SMGLoM exists) and a starting symbol $s\in\mathtt{Sym}$ (e.g. `nattimes`; binary multiplication on natural numbers). The algorithm then proceeds as follows:

\begin{enumerate}
	\item If $s:T$ has a (simple or dependent) function type, we fill in the required arguments. For $s=$`nattimes`, our type is $T=$`Nat`$\to$`Nat`$\to$`Nat`, hence we need to find two arguments $s_1,s_2$ of type `Nat`. For each $s_i$ of required type $T_i$ we proceed as follows:
		\begin{enumerate}
			\item With probability $p_{var}$, we introduce a new variable $v:T_i$ from a list of allowed variable names (which include variants such as $a$, $a'$, $a_0$ etc.) and let $s_i:=v$.
			\item With probability $p_{fun}$, we pick a symbol $f\in \mathtt{Sym}$ with a function type with return type $T_i$ (e.g. for $T_i=$`Nat`, we can pick `natplus`). In that case, we let $s:=f$, recurse, and set $s_i$ as the result.
			\item With probability $p_{const}=1-p_{var}-p_{fun}$, we pick a constant symbol $c\in \mathtt{Sym}$ of type $T_i$ (e.g. for $T_i=$`Nat` we can pick $0$) and return $s_i:=c$.
		\end{enumerate}
		In order to avoid stack overflows, we reduce $p_{fun}$ in each iteration by a certain factor $<1$. As to not overuse certain symbols, we scale $p_{fun}$ and $p_{const}$ with the number of respectively suitable symbols available; if $\mathtt{Sym}$ contains no suitable function or constant symbols, we let $p_{fun}=0$ (and/or $p_{const}=0$, respectively).
	\item If $s:T$ does \emph{not} have a function type (or all its parameters have been filled in 1.), then $s$ is well-typed and we return $s$ with probability $1-p_{up}$. 
	
	With probability $p_{up}$, we instead pick a new symbol $s_f\in S$ of some function type such that some $i$-th parameter type of $s_f$ is $T$. In that case, we let $s_i:=s$ and $s:=s_f$ and recurse.
	
	Again, in order to avoid stack overflows we reduce $p_{up}$ by some factor with each iteration.
\end{enumerate}

The algorithm also takes subtyping into account, e.g. whenever a term of type `Real` is required, terms of type `Int` or `Nat` are used with some probability.

\paragraph{} In order to obtain a sentence in the sense of \autoref{sec:datasets} providing context for disambiguation, we first translate $t$ along alignments to SMGLoM (using a random `\symvariant`), collect the set $V$ of all free variables of $t$ and \emph{verbalize} their types. For that, we associate each \emph{type} with a set of \emph{verbalizations} from which we choose randomly to produce a sentence that introduces the variables before using them in the generated expression. \autoref{ex:verbalization} shows a few example verbalizations for a variable $x$ of type `Nat` and generated sentences for the input symbol $s=$`realuminus`; the negation on real numbers. 

\ifnips\begin{figure}[ht]\else\begin{figure*}[ht]\fi
\begin{center}
\begin{tabular}{|r||l|l|}\hline
	& Generated \sTeX & PDF output \\\hline\hline
Verbalizations & `$\inset{x}{\NaturalNumbers}$` & $x\in\mathbb N$ \\
	& a positive integer `$x$` & a positive integer $x$ \\
	& an integer `$\intmethan{x}{0}$` & an integer $x\geq0$ \\
	& a natural number `$x$` & a natural number $x$ \\\hline
Sentences &
	\begin{minipage}[t]{0.47\textwidth}
		`Assume we have some $\inset{y'}` \\
		`{\NaturalNumbers}$ and arbitrary ` \\
		`$\inset{\mathcal F}{\IntegerNumbers}$.` \\ 
		`It follows that $\realuminus{\realuminus` \\
		`{\inttimes[x]{\mathcal F,y',y'}}}$.`
	\end{minipage} & \begin{minipage}[t]{0.3\textwidth}
		Assume we have some $y'\in \mathbb N$ and arbitrary $\mathcal F\in\mathbb Z$.
		It follows that $--(\mathcal F\times y'\times y')$.
	\end{minipage} \\ &
	\begin{minipage}[t]{0.47\textwidth}
	\rule{\textwidth}{1pt}\\
		`Let $\natmorethan n{0}$. Then consider` \\
		`$\realuminus{\realuminus{\natsucc{` \\
		`\natsucc n}}}$.`
	\end{minipage} & \begin{minipage}[t]{0.3\textwidth}
	\rule{\textwidth}{1pt}\\
		Let $n>0$. Then consider\\ $--S(S(n))$.
	\end{minipage}\\ &
	\begin{minipage}[t]{0.47\textwidth}
	\rule{\textwidth}{1pt}\\
		`Whenever we have some positive natural` \\
		`number $\varepsilon$, any integer $\ell$` \\ 
		`and a real number $\livar{\mathcal C}` \\
		`{2}$, then it follows that $\realtimes{` \\
		`\livar{\mathcal C}{2},\livar{\mathcal C}`\\
		`{2},\realplus{\realuminus{\ell},\natsucc{` \\
		`\varepsilon}}}$.`
	\end{minipage} & \begin{minipage}[t]{0.3\textwidth}
	\rule{\textwidth}{1pt}\\
		Whenever we have some positive natural number $\varepsilon$, any integer $\ell$ and a real number $\mathcal C_2$, then it follows that $\mathcal C_2 \mathcal C_2(-\ell + S(\varepsilon))$.
	\end{minipage}\\\hline
\end{tabular}
\end{center}
\caption{Example Verbalizations for $x:$Nat and Generated Sentences}\label{ex:verbalization}
\ifnips\end{figure}\else\end{figure*}\fi

The verbalizations are categorized as \emph{prefixed} (e.g. \lquot \emph{a natural number $n$}'') or \emph{suffixed} (e.g. \lquot \emph{$n$ a natural number}''), and \emph{singular} or \emph{plural}, and picked according to the number of variables of the same type and the surrounding sentence, which is also picked at random (e.g. \lquot\emph{Assume we have ...}'' uses prefixed, whereas \lquot\emph{Let ...}'' uses suffixed).  

\section{Evaluation Tactics}\label{apx:evaluation}

For every \LaTeX\ input $S_\mathlatex$, expected label $S_\mathstex$ and returned sentence $S_R$, 
we employ the following strategies, the results of which are summarized in \autoref{fig:eval}:
\begin{itemize}
	\item[\texttt{islatex}] We parse $S_R$ into an AST. Success implies that $S_R$ is syntactically valid \LaTeX. This might fail for \lquot minor'' reasons such as a missing closing bracket. It might yield false positives in cases where macros (not explicitly considered by our parser) occurring in $S_R$ have a wrong number of arguments.
	
	All subsequent evaluation strategies require \texttt{islatex} to succeed.
	\item[\texttt{stexcheck}] We heuristically check whether $S_R$ is in $\mathcal L_\mathstex$ -- unlike \texttt{islatex}, this requires that all \sTeX macros occurring in $S_R$ have the right number of arguments. Success does \emph{not} tell us that the input has been disambiguated \emph{correctly}, but \emph{does} imply that is \emph{has} been disambiguated \emph{at all}. False negatives can occur if $S_R$ (and thus likely $S_\mathlatex$ as well) contains complex variable names, or if $S_R$ contains e.g. an equality symbol \lquot ='' instead of the corresponding \sTeX macro, which LaTeXML could recover.
	
	\item[\texttt{eval\_latex}] All \sTeX macros occurring in $S_R$ are expanded and $S_R$ is normalized as described in \autoref{sec:datasets}. The result is string-compared to $S_\mathlatex$. Success thus implies, that the notational presentation in PDF output of $S_\mathlatex$ and $S_R$ will coincide. False negatives can occur due to minor differences e.g. in not strictly necessary brackets.
	
	\item[\texttt{omdoc}] $S_R$ is translated to \omdoc using LaTeXML and imported to \mmt. Success guarantees syntactic well-formedness of $S_R$. Since both the LaTeXML-\omdoc export and the subsequent \mmt-import are somewhat brittle, this can easily lead to false negatives.
	
	\item[\texttt{translated}] The import from \texttt{omdoc} is translated to the typed MitM library. This entails that all symbols used in $S_R$ are aligned with MitM symbols and $S_R$ is amenable for formal knowledge management services.
	
	\item[\texttt{inferred}] The translation to MitM obtained from \texttt{translated} is type checked by \mmt by having its type inferred. Success guarantees that $S_R$ is well-typed.
	
	Notably, if $S_R$ is a mere variable (e.g. the expression `$n$`), it does not actually have an inferrable type, but succeeds trivially. This accounts for 60 of the entries in our evaluation set, i.e. 37\%.
	
	\item[\texttt{provided\_stex}] Both the expected label $S_\mathstex$ and $S_R$ are normalized and string-compared. Success implies that $S_R$ is definitely the correct translation. False negatives can easily occur due to non-semantic differences between $S_\mathstex$ and $S_R$ however, such as bracketing, nested applications in $S_R$ (e.g. `$\natplus{\natplus{a,b},c}$` vs. `$\natplus{a,b,c}$`), etc.
	
	\item[\texttt{stex\_as\_omdoc}] $S_\mathstex$ is translated to \omdoc via LaTeXML and directly compared to the \omdoc-term obtained from \texttt{omdoc}. Like \texttt{provided\_stex}, success implies that $S_R$ is correct, but it is more fault-tolerant with respect to the precise syntax of $S_R$, while being \emph{less} fault tolerant due to the issues mentioned in \texttt{omdoc}.
\end{itemize}

\begin{figure}[ht]
\begin{center}\begin{tabular}{|r||l|}\hline
	\emph{Total inputs} & 161 \\\hline\hline
	\texttt{islatex} & 96.9\% \\
	\texttt{stexcheck} & 60.2\% \\
	\texttt{eval\_latex} & 64.0 \% \\\hline
	\texttt{omdoc} & 76.4\% \\
	\texttt{translated} & 63.5\% \\
	\texttt{inferred} & 59.6\% \\\hline
	\texttt{provided\_stex} & 47.2 \% \\
	\texttt{stex\_as\_omdoc} & 53.4 \% \\\hline
\end{tabular}
\end{center}

\caption{Results on our Evaluation Document}\label{fig:eval}
\end{figure}

The first three evaluations can always be applied; from the remaining,
all but \texttt{provided\_stex} require a working installation of LaTeXML and its \sTeX-Plugin. The last two require a known correct translation.

A detailed log file on our evaluation document with the individual results for each input and evaluation is available in the associated git repository%\footnote{\url{https://gl.kwarc.info/dmueller/fifom/-/blob/master/deploy/test.log}}
.

\lstDeleteShortInline`